\documentclass[11pt,a4paper]{article}
\usepackage[nohyperref]{acl2017}
\usepackage{times}
\usepackage{latexsym}
\usepackage{amssymb}
\usepackage{amsmath}

\usepackage{url}

\usepackage{graphicx}
\usepackage{tabulary}

\aclfinalcopy % Uncomment this line for the final submission
\title{Multiple Range-Restricted Bidirectional Gated Recurrent Units with Attention for Relation Classification}

\author{Jonggu Kim \\
  Computer Science and Engineering, \\
  Pohang University of Science and \\ Technology (POSTECH) \\
  Pohang, Republic of Korea \\
  {\tt jgkimi@postech.ac.kr} \\\And
  Jong-Hyeok Lee \\
  Computer Science and Engineering, \\
  Pohang University of Science and \\ Technology (POSTECH) \\
  Pohang, Republic of Korea \\
  {\tt jhlee@postech.ac.kr} \\}

\date{4th, Feb., 2017}

\begin{document}
\maketitle
\begin{abstract}
 Most of neural approaches to relation classification have focused on finding short patterns that represent the semantic relation using Convolutional Neural Networks (CNNs) and those approaches have generally achieved better performances than using Recurrent Neural Networks (RNNs). In a similar intuition to the CNN models, we propose a novel RNN-based model that strongly focuses on only important parts of a sentence using multiple range-restricted bidirectional layers and attention for relation classification. Experimental results on the SemEval-2010 relation classification task show that our model is comparable to the state-of-the-art CNN-based and RNN-based models that use additional linguistic information.
\end{abstract}

\section{Introduction}

Relation classification is to select the relation class that implies the relation of the two nominals (e1, e2) in the given text. For instance, given the following sentence, ``The \textless e1\textgreater \textbf{phone}\textless /e1\textgreater{} went into the \textless e2\textgreater \textbf{washer}\textless /e2\textgreater{}.'', where \textless e1\textgreater{}, \textless /e1\textgreater{}, \textless e2\textgreater{}, \textless /e2\textgreater{} are position indicators that represent the starting and ending positions of nominals, the goal is to find the actual relation \textit{Entity-Destination} of \textbf{phone} and \textbf{washer}. The task is important because the results can be utilized in other Natural Language Processing (NLP) applications like question answering and information retrieval.

Recently, Neural Network (NN) approaches to relation classification have been spotlighted since they do not need any handcrafted features but even obtain better performances than traditional models. Such NNs can be simply classified into CNN-based and RNN-based models, and they capture slightly different features to predict a relation class.

In general, CNN-based models can only capture local features while RNN-based models are expected to capture global features as well, but the performances of CNN-based models are better than RNN-based models. That could be thought that most of relation-related terms are not scattered but intensively positioned as short expressions on a given sentence, and further even if RNNs are expected to learn such information automatically, it cannot be easily done contrary to our expectation. To overcome the limitation of RNNs, most of the recent work using RNNs have used additional linguistic information like Shortest Dependency Path (SDP), which can reduce the effect of noise words when predicting a relation.

In this paper, we propose a simple RNN-based model that strongly pays attention to nominal-related and relation-related parts with multiple range-restricted RNN variants called Gated Recurrent Units (GRUs) \cite{Cho:14} and attention. On the SemEval-2010 Task 8 dataset \cite{Hendrickx:09}, our model with only pretrained word embeddings achieved the F1 score of 84.3\%{}, which is comparable with the state-of-the-art CNN-based and RNN-based models that use additional linguistic resources such as Part-Of-Speech (POS) tags, WordNet and SDP. Our contributions are summarized as follows:
\begin{itemize}
\item For relation classification, without any additional linguistic information, we suggest modeling nominals and a relation in a sentence with specified range-restriction standards and attention using RNNs.
\item We show how effective abstracting nominal parts, a relation part and both separately with the restrictions is to relation classification.
\end{itemize}

\section{Related Work}

Traditional approaches to relation classification are to find important features of relations with various linguistic processors and utilize them to train classifiers. For instance, \newcite{Rink:10} uses NLP tools to extract linguistic features and trains an SVM model with the features.

Recently, many deep learning approaches have been proposed. \newcite{Zeng:14} proposes a model based on CNNs to automatically learn important N-gram features. \newcite{Santos:15} proposes a ranking loss function to well distinguish between the real classes and \textit{Other} class. To capture long distance patterns, RNN-based, usually using Long Short-Term Memory (LSTM), approaches have also appeared, one of which is \newcite{Zhang:15}. The model simply feeds on all words in a sentence, then captures important one through the max-pooling operation. \newcite{Xu:15} and \newcite{Miwa:16} propose other RNN models using SDP to ignore noise words in a sentence. In addition, \newcite{Liu:15} and \newcite{Cai:16} propose hybrid models of RNN and CNN.

One of the most related work to ours is the attention-based bidirectional LSTM (att-BLSTM) \cite{Zhou:16}. The model uses bidirectional LSTM and attention techniques to abstract important parts. However, the att-BLSTM does not distinguish roles of each part in a sentence, which could not involve sensitive attention. Another of the most related work is by \newcite{Zheng:16}. They try to capture nominal-related and relation-related patterns with CNNs and use neither restrictions nor attention mechanism.

\section{The Proposed Model}
\begin{figure*}[t]
  \centering
  \noindent
  \includegraphics[width=\linewidth]{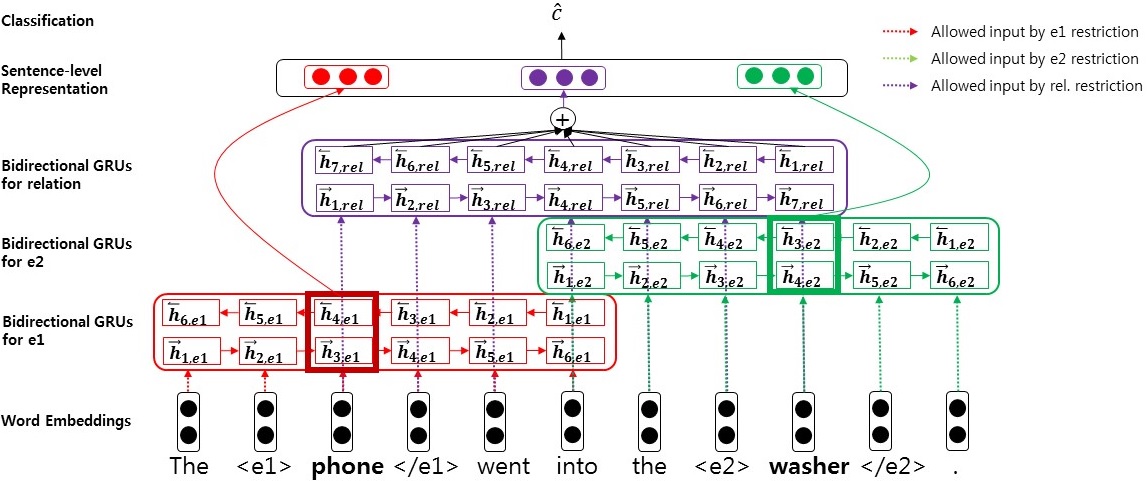}
  \caption{Multiple Range-Restricted Bidirectional GRUs with Attention (\(k = 3\))}
  \label{fig:architecture}
\end{figure*}

Figure \ref{fig:architecture} shows the architecture of the proposed model, which will be described in the subsections.

\subsection{Word Embeddings}

Our model first takes word embeddings to represent a sentence at the word level.
Given a sentence \(S\) consisting of \(N\) words, it can be represented as \(S = \{w_1\) , \(w_2\), \(w_3\),..., \(w_N\)\}. We convert each one-hot vector \(w_t\) by multiplying with the word embedding matrix \( W_{e} \in \mathbb{R}^{d_{e} \times |V|}\):

\begin{equation}
 e_t = W_{e} w_t .
\end{equation}

Then, the sentence can be represented as \(S_e = \{e_1, e_2,..., e_N\}\).

\subsection{Range-Restricted Bidirectional GRUs}
To capture information of two nominals and one relation, our model consists of three bidirectional GRU layers with range restrictions. A GRU is a kind of RNN variant to alleviate the gradient-vanishing problem like LSTM, but it has fewer weights than LSTM. In a GRU, the \(t\)-th hidden \(h_t\) with reset gate \(r_t\) and update gate \(z_t\) is computed as:

\begin{flalign}
 & r_t = \sigma \big(W_r e_t + U_r h_{t-1}\big) , \\
 & z_t = \sigma \big(W_z e_t + U_z h_{t-1}\big), \\
 & \tilde{h}_t = tanh \big(W e_t + U (r_t \odot h_{t-1})\big) , \\
 & h_t = z_t \odot h_{t-1} + (1-z_t) \odot \tilde{h}_t ,
\end{flalign}
where \(\sigma\) is the logistic sigmoid function.

The range restrictions can be done by using masking techniques to restrict the input range of the three bidirectional GRUs. Therefore, they should be conducted under three separate standards, but because the standards for two nominals are the same, we introduce two kinds of standards. First, to capture  each nominal information, only the \( p_{en} \pm k \) positioned words are regarded as input to the corresponding bidirectional GRU layer, where \( p_{en}\) is the position of nominal e1 or e2 and \( k \) is a hyperparameter affecting their window size. Second, for the relation GRU layer, the input range is set to \([p_{e1} , p_{e2}]\) or \([p_{e2} , p_{e1}]\) according to the relative order of the nominals in a sentence, which means that the range is from the formerly-appearing nominal to the latterly-appearing nominal.

After the sentence representation at word level \(S_e\) is fed into the six GRU layers (three GRU layers in two directions) under the restrictions, various hidden units are finally generated from the layers. We call the hidden units of each GRU layer \(\overrightarrow{H}_{e1}, \overleftarrow{H}_{e1}, \overrightarrow{H}_{e2}, \overleftarrow{H}_{e2}, \overrightarrow{H}_{rel}, \overleftarrow{H}_{rel}\) for convenience in the next subsection.

\subsection{Sentence-level Representation}
Among the hidden units of the six range-restricted GRUs, the model selects important parts by using direct selection from hidden layers and the attention mechanism.

To extract e1 and e2 information, we propose to directly select hidden units at each nominal position in the e1 and e2 bidirectional GRUs, and to sum them to construct \(v_{e1}\), \(v_{e2} \in \mathbb{R}^{d_{h}}\), respectively as:
\begin{flalign}
  & v_{e1} = \overrightarrow{h}_{e1} + \overleftarrow{h}_{e1} , \\
  & v_{e2} = \overrightarrow{h}_{e2} + \overleftarrow{h}_{e2} ,
\end{flalign}
where each directional \( h_{en} \) represents hidden units at the \(en\) positions in the directional \(H_{en}\).

To abstract relation information, we adopt the attention mechanism that has been widely used in many areas \cite{Bahdanau:14,Hermann:15,Chorowski:15,Xu:15att}. We use the attention mechanism \cite{Zhou:16}, but we apply it to each directional GRU layer independently to capture more informative parts with the flexibility. The forward directional relation-abstracted vector \( \overrightarrow{v}_{rel} \) is computed as (\( \overleftarrow{v}_{rel} \) in the same way):
\begin{flalign}
  & \overrightarrow{M} = tanh( \overrightarrow{H}_{rel} ) ,\\
  & \overrightarrow{\alpha} = softmax( \overrightarrow{w}_{att}^T \overrightarrow{M} ) ,\\
  & \overrightarrow{v}_{rel} = \overrightarrow{H}_{rel} \overrightarrow{\alpha}^T ,
\end{flalign}
where \(\overrightarrow{w}_{att}\) is a trained attention vector for the forward layer.

Then, we sum \( \overrightarrow{v}_{rel} \) and \( \overleftarrow{v}_{rel} \) to make the relation-abstracted vector \(v_{rel} \in \mathbb{R}^{d_{h}}\):
\begin{equation}
  v_{rel} = \overrightarrow{v}_{rel} + \overleftarrow{v}_{rel} .
\end{equation}

Lastly, the final representation \(v_{fin} \in \mathbb{R}^{3d_{h}}\) is constructed by concatenating them:
\begin{equation}
  v_{fin} = v_{e1} \oplus v_{rel} \oplus v_{e2} ,
\end{equation}
where \(\oplus\)  is a concatenation operator.

\newcolumntype{L}[1]{>{\raggedright\let\newline\\\arraybackslash\hspace{0pt}}m{#1}}
\newcolumntype{C}[1]{>{\centering\let\newline\\\arraybackslash\hspace{0pt}}m{#1}}
\newcolumntype{R}[1]{>{\raggedleft\let\newline\\\arraybackslash\hspace{0pt}}m{#1}}

\begin{table*}[t]
  \centering
  \begin{tabulary}{\textwidth}{ |C{5.6cm}|L{8.3cm}|C{0.7cm}| }
    \hline
    \textbf{Model} & \textbf{Additional Features (Except Word Embeddings)} & \textbf{F1} \\
    \hline
    SDP-LSTM \newline \cite{Xu:15} & - POS, WordNet, dependency parse, grammar relation & 83.7 \\ 
    \hline
    DepNN \newline \cite{Liu:15} & - NER, dependency parse & 83.6 \\ 
    \hline
    SPTree \newline \cite{Miwa:16} & - POS, dependency parse & 84.4 \\ 
    \hline
    MixCNN+CNN \newline \cite{Zheng:16} & - None & \textbf{84.8} \\ 
    \hline
    att-BLSTM \newline \cite{Zhou:16} & - None & 84.0 \\ 
    \hline
    Our Model (att-BGRU) \newline Our Model (Relation only) \newline Our Model (Nominals only) \newline Our Model (Nominals and Relation) & - None \newline - None \newline - None \newline - None & 82.9 \newline 83.0  \newline 81.4 \newline \textbf{84.3} \\
    \hline
  \end{tabulary}
  \caption{Comparison with the results of the state-of-the-art models}
  \label{table:perf}
\end{table*}

\subsection{Classification}
Our model uses scores of how similar the \(v_{fin}\) is to each class embedding to predict the actual relation \cite{Santos:15}. Concretely, we propose a feed-forward layer in which a weight matrix \(W_{c} \in \mathbb{R}^{classes \times 3d_{h}} \) and a bias vector \(b_c \in \mathbb{R}^{classes}\) can be regarded as a set of the class embeddings. In other words, the inner-product of each row vector in \(W_{c}\) with \(v_{fin}\) represents the similarity between them in vector space, so the class score vector \(s_c \in \mathbb{R}^{classes} \) is just computed as:
\begin{equation}
   s_c = W_{c} v_{fin} + b_{c} .
\end{equation}

Then, the model chooses the max-valued index that represents the most probable class label \(\hat{c}\) except that every value in the \(s_c\) is negative. In the exceptional case, \(\hat{c}\) is chosen as \textit{Other} \cite{Santos:15}.

\subsection{Training Objectives}

We adopt the ranking loss function \cite{Santos:15} to train the networks. Let \(s_{cy^+}\) the score of the \(\hat{c}\), and \(s_{cc^-}\) the competitive score that is the best score excluding \(s_{cy^+}\) for convenience. Then, the loss is computed as:
\begin{multline}
  L = log(1 + exp(\gamma(m^+ - s_{cy^+})))\\
     + log(1 + exp(\gamma(m^- + s_{cc^-}))) ,
\end{multline}

where \(m^+\) and \(m^-\) represent margins and \(\gamma\) is a factor that magnifies the gap between the score and the margin.

\section{Experiments}

For the experiments, we implement our model in Python using Theano \cite{theano:16} and use the model with the following descriptions.

\subsection{Datasets and Settings}
We conduct the experiments with  SemEval-2010 Task 8 dataset \cite{Hendrickx:09}, which contains 8,000 sentences as the training dataset, and 2,717 sentences as the test dataset. A sentence consists of two nominals (e1, e2), and a relation between them. Ten relation types are considered: Nine specific types (\textit{Cause-Effect, Component-Whole, Content-Container, Entity-Destination, Entity-Origin, Instrument-Agency, Member-Collection, Message-Topic and Product-Producer}), and the \textit{Other} class. The specific types have directionality, so a total of \(2\times9+1 = 19\) relation classes exist.

We use 10-fold cross-validation to tune the hyperparameters. We adopt the 100-dimensional word vectors trained by \newcite{Pennington:14} as initial word embeddings and select the hidden layer dimension \(d_n\) of 100, the learning rate of 1.0 and the batch size of 10. AdaDelta \cite{Zeiler:12} is used as the learning optimizer. Also, we adapt the dropout \cite{Hinton:12} to the word embeddings, GRU hidden units, and feed-forward layer with dropout rates of 0.3, 0.3 and 0.7, respectively, and use the \(k\) of 3. We adopt the position indicator that regards \textless{}e1\textgreater, \textless{}/e1\textgreater, \textless{}e2\textgreater{} and \textless{}/e2\textgreater{} as single words \cite{Zhang:15}. We set \(m^+\), \(m^-\) and \(\gamma\) to 2.5, 0.5 and 2.0, respectively \cite{Santos:15} and adopt the L2 regularization with \(10^{-5}\). The official scorer is used to evaluate our model in the macro-averaged F1 (excluding \textit{Other}).

\subsection{Results}
In Table \ref{table:perf}, our results are compared with the other state-the-art models. Our model with only pretrained word embeddings achieved the F1 score of 84.3\%, which is comparable to the state-of-the-art models.

Furthermore, we investigated the effects of extracting relation, nominals and both of them. Attention-based bidirectional GRUs with no restriction (att-BGRU) were also tested as a reimplementation of the att-BLSTM. Here, our finding is that the restricted version of the att-BGRU (the relation only model) is not significantly better, but by abstracting nominals together, the model achieves higher F1 score. That indicates even if the ranges are slightly overlapped, they capture distinct features and improve the performance.

\section{Conclusion}

This paper proposed a novel model based on multiple range-restricted RNNs with attention. The proposed model achieved a comparable performance to the state-of-the-art models without any additional linguistic information.

%\bibliographystyle{acl_natbib_mine}
%\bibliography{acl2017_mine}
\bibliography{acl2017_mine}
\bibliographystyle{acl_natbib_mine}

\end{document}